\documentclass[letterpaper, 10 pt, conference]{ieeeconf}  

\IEEEoverridecommandlockouts                              

\makeatletter
\let\NAT@parse\undefined
\makeatother
\usepackage[numbers,sort&compress]{natbib}

\usepackage[pdftex]{graphicx}
\usepackage{amsmath}
\usepackage{amssymb}  
\usepackage{subfigure}
\usepackage{multirow}
\usepackage{array,booktabs}
\usepackage{diagbox}
\usepackage{balance}
\usepackage{xspace}
\usepackage{algorithm}
\usepackage{algpseudocode}
\usepackage{adjustbox}
\usepackage{romannum}

 \usepackage[final]{hyperref}
 \hypersetup{
     colorlinks=true,
     linkcolor=blue,
     filecolor=magenta,
     urlcolor=blue,
     citecolor=black
 }
\usepackage[font=small]{caption}
\usepackage{rpm_SIunits}
\usepackage{rpm_acronyms}
\usepackage{rpm_math}
\usepackage{rpm_misc}

\usepackage{soul,color}
\usepackage{lipsum}

\usepackage{xcolor}

\usepackage{tikz} 
\usepackage{subcaption}
\usepackage{caption}
\title{\LARGE \bf PeLiCal: Targetless Extrinsic Calibration via Penetrating \\ Lines for RGB-D Cameras with Limited Co-visibility}     

\author{Jaeho Shin${}^{1}$, Seungsang Yun${}^{1}$ and Ayoung Kim${}^{1*}$
\thanks{$^\dagger$This work was supported by IITP (No.2022-0-00480), KIAT (P0020536) and Interdisciplinary Research Initiatives Program SNU (2023).}
\thanks{$^{1}$J. Shin, S. Yun and A. Kim are with the Dept. of Mechanical Engineering, SNU, Seoul, S. Korea {\tt\small [leah100, seungsang, ayoungk]@snu.ac.kr}}%
}
\begin{document}

\maketitle
\thispagestyle{empty}
\pagestyle{empty}

\begin{abstract}
RGB-D cameras are crucial in robotic perception, given their ability to produce images augmented with depth data. However, their limited \ac{FOV} often requires multiple cameras to cover a broader area. In multi-camera RGB-D setups, the goal is typically to reduce camera overlap, optimizing spatial coverage with as few cameras as possible. The extrinsic calibration of these systems introduces additional complexities. Existing methods for extrinsic calibration either necessitate specific tools or highly depend on the accuracy of camera motion estimation. To address these issues, we present PeLiCal, a novel line-based calibration approach for RGB-D camera systems exhibiting limited overlap. Our method leverages long line features from surroundings, and filters out outliers with a novel convergence voting algorithm, achieving targetless, real-time, and outlier-robust performance compared to existing methods. We open source our implementation on \url{https://github.com/joomeok/PeLiCal.git}.
\end{abstract}
\section{Introduction}
\label{sec:intro}
In robotic perception, RGB-D cameras serve as pivotal sensors due to their capacity to capture images accompanied by corresponding depth values. Due to the limited \ac{FOV} of the sensor, common practice is to employ a multi-camera configuration to ensure comprehensive coverage of the surrounding environment.


In systems deploying multiple RGB-D cameras, unlike RGB camera setups designed for shared \ac{FOV} (e.g., binocular camera configurations), the system is set up to reduce the overlap in the \ac{FOV}. This strategy aims to cover the nearby environment using a minimum number of cameras. However, such configuration makes it challenging to employ a calibration pattern observed simultaneously within the shared \ac{FOV} when calibrating the extrinsic parameters of the system. Several approaches have been proposed to address these challenges in the extrinsic calibration of camera systems with limited co-visibility.

Existing works bifurcate into two distinct approaches: one leveraging specialized equipment and the other adopting a hand-eye-like calibration by estimating each camera motion~\cite{Pattern-2007-Esquivel}. In the former case, \citeauthor{IROS-2013-Li}~\cite{IROS-2013-Li} utilized a target with unique features to easily estimate the location of points in the partially detected pattern board. In the other case, \citeauthor{CVPR-2008-Kumar}~\cite{CVPR-2008-Kumar} introduced additional apparatuses like mirrors. Unfortunately, these approaches are impractical because employing such specialized tools for calibration is burdensome. In the latter case, the calibration results are highly dependent on the accuracy of the estimated camera motion.

\begin{figure}[!t]
    \centering
    \includegraphics[width=1\columnwidth]{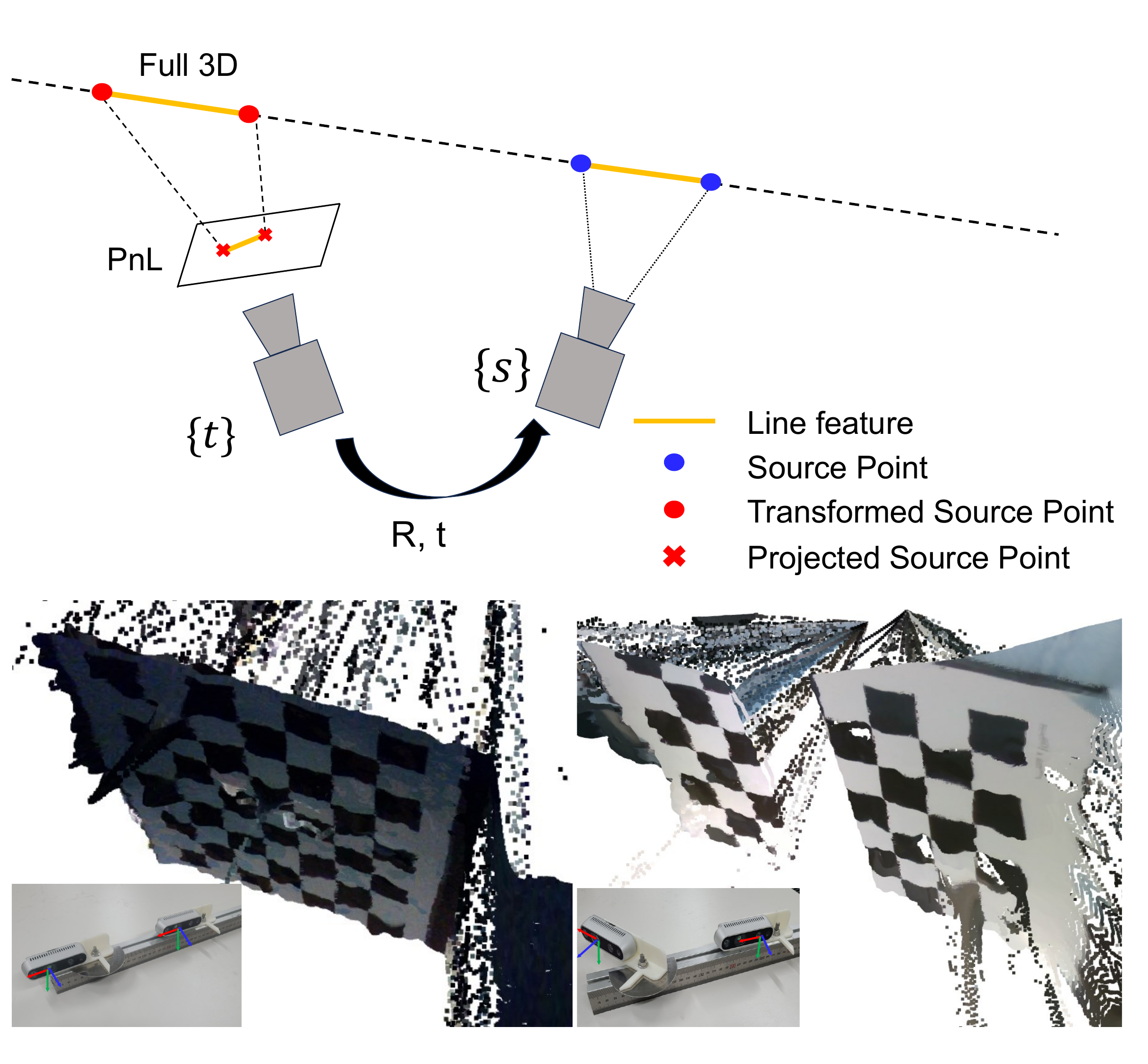}
    \caption{
According to the inlier ratio of RANSAC-based fitting from the depth images, each line pair is classified into a \textit{full 3D} case or \textit{\ac{PnL}} case, and different constraints are employed (top). Two checkerboard planes from each camera are merged by extrinsic parameters estimated from PeLiCal (bottom). Our method uses long lines matched from both images, and on the condition that the cameras are set up to observe the same feature, it can reliably determine the pose in cases with sufficient overlap (left) as well as in cases without any overlap (right). 
    }
    \label{fig:overview}
    \vspace{-3mm}
\end{figure}

Overcoming these challenges, we introduce a targetless but robust methodology for extrinsic calibration, specifically for RGB-D camera systems with limited overlap. Without requiring a large calibration target, which is commonly observable, we focus on line features intersecting two cameras. Accordingly, our algorithm performs reliably in real-time, without a pattern board, additional external devices, or inter-frame motion estimation, even if co-visibility is limited. Additionally, measurements that induce estimations closely matching the conditions of the rotation matrix and which are confirmed as convergent by a voting algorithm are incorporated, enhancing robustness against outliers. The key contributions of our study are encapsulated as follows:

\begin{itemize}
    \item We present a robust, targetless approach for calibrating the extrinsic parameters of RGB-D cameras across diverse \ac{FOV} settings using penetrating line matching, available from surroundings.
    
    \item Our algorithm selectively incorporates informative scenes by projecting the estimated rotation matrix on the $SO(3)$ manifold and calculating its shortest distance. Furthermore, the existence of precise translation is validated by examining the convergence of 3D geometric constraints, derived from the transformation of Plücker coordinates.
    
    \item The calibration accuracy of our method is validated from various configurations of \ac{FOV} using a specially designed device. The algorithm shows superior and stable accuracy with existing calibration tools, especially in settings with an increased baseline in stereo setups.
\end{itemize}

\section{related work}
\label{sec:relatedwork}
\subsection{Line Based Camera Pose Estimation}
Leveraging line features has been widely adopted to estimate the camera pose. Among these approaches, the algorithm that relies on matching 3D to 2D pairs is the \ac{PnL}. Most studies estimate the camera pose by formulating equations based on the associated 3D and 2D line features. \citeauthor{DARPA-1993-Hartley} in \cite{DARPA-1993-Hartley} proposed the method to calculate the essential matrix from a set of nine lines. In the context of associating 3D to 2D line pairs, a commonly invoked constraint is that the preimage of the 2D line should encapsulate the transformed 3D line as in \cite{ICRA-2011-Mirzaei}. \citeauthor{Arxiv-2016-Pribyl} employ representation of lines using plücker coordinates to estimate the camera pose in \cite{Arxiv-2016-Pribyl}. \citeauthor{TPAMI-2016-Xu} outlined the computation of the rotation matrix by decomposing it into two distinct rotation angles incorporating an auxiliary model frame, as detailed in \cite{TPAMI-2016-Xu}. The geometric condition of the line presented in the previous research is used as a theoretical background for our method.

\subsection{Non-overlapping FOV Camera Calibration}
In a study where the \ac{FOV} of the cameras exhibit no overlap, the methodology proposed by \citeauthor{CVPR-2008-Kumar} in \cite{CVPR-2008-Kumar} employs reflected images of a chessboard via a planar mirror to determine extrinsic parameters. In \cite{IROS-2013-Li}, \citeauthor{IROS-2013-Li} proposes a calibration method using a calibration pattern enriched with unique features, which enables accurate localization of features, even when the pattern board is partially observed. However, these techniques are impractical because specialized equipment or a substantially large pattern board is required. Several algorithms have been developed to alleviate this challenge that harnesses camera motion estimation to retrieve the 6-DOF extrinsic parameters as in \cite{Pattern-2007-Esquivel, 3DV-2016-Zhu}. However, these motion-centric calibration approaches are intrinsically reliant on the precision of the camera's motion. 

The investigation conducted by \citeauthor{RAL-2017-Perez} closely corresponds to our algorithm since it exploits 3D lines to calibrate the limited co-visibility cameras in \cite{RAL-2017-Perez}. Their method entails extracting 2D and 3D lines and projecting 3D lines onto 2D images using initial extrinsic parameters to establish correspondences among line features. Subsequently, the extrinsic parameters are refined repeatedly by matching the feature based on the estimated pose until convergence. Given the dependence on the initial parameter for associating line features, achieving accurate calibration parameters is infeasible when the initial value is erroneous. In contrast, our proposed algorithm calculates initial parameters using robustly matched lines, followed by feasibility evaluation of the solution and optimization. As a result, consistent calibration parameters can be obtained regardless of the initial conditions.
\section{Methodology}
\label{sec:method}

\subsection{Notation}
\begin{figure}[t!]
    \centering
    \includegraphics[width=1\columnwidth]{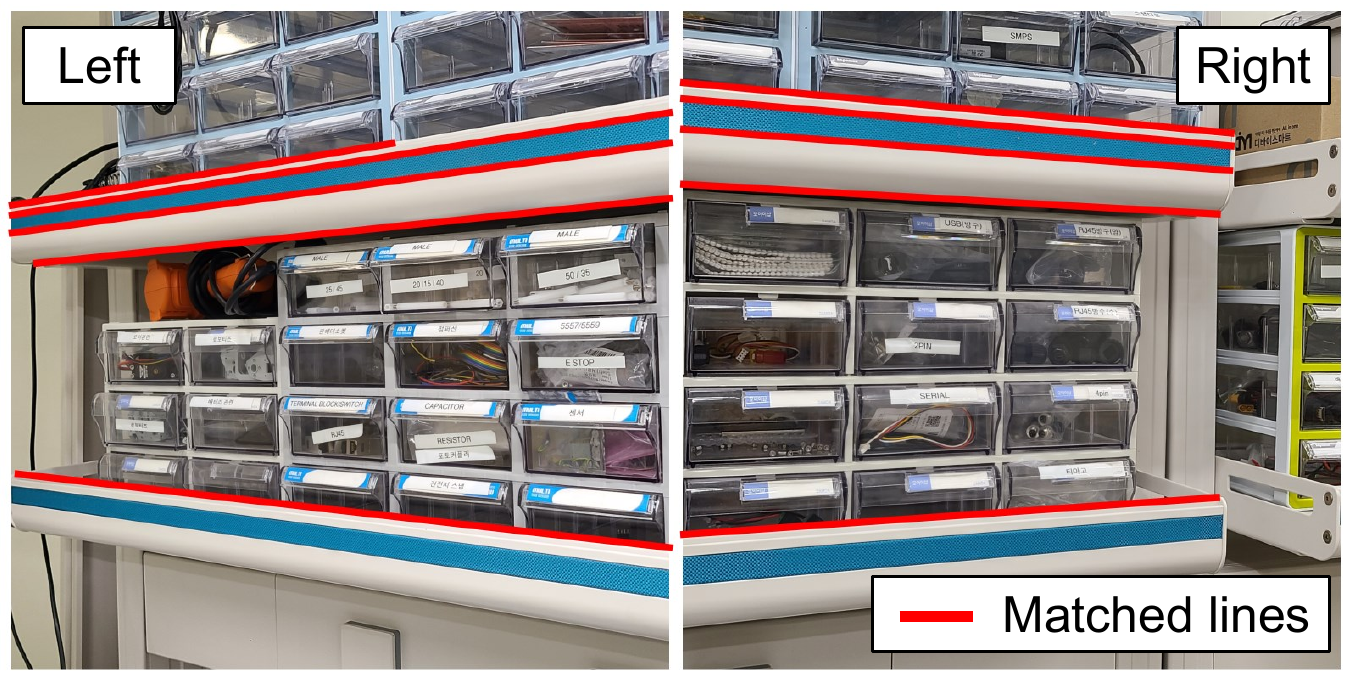}
    \caption{ Line features (depicted in red) can be matched between two cameras even when they lack overlapping FOV areas. In contrast to point features, long lines can identify corresponding pairs under these conditions, allowing for determining geometric constraints.
    }
    \label{fig:matching}
    \vspace{-0.5cm}
\end{figure}
In accordance with our notation conventions, we represent scalars using italicized characters, vectors with boldface lowercase characters, and matrices with boldface uppercase characters. The subscripts $t$ and $s$ are utilized to denote parameters associated with the target and source cameras, respectively. For clarity in notation, a 3D line is represented as $\boldsymbol{\mathrm{L}}$, with its corresponding Plücker coordinates expressed as $\boldsymbol{\mathcal{L}} = [\boldsymbol{\mathrm{d}}, \boldsymbol{\mathrm{m}}]$. Correspondingly, its associated dual Plücker matrix is represented as $\boldsymbol{\mathrm{L}}^*$. Within the 2D image plane, the notation $\mathbf{l}$ represents a 2D line. The matrix $\boldsymbol{\mathcal{K}}$ denotes the line projection matrix, while $\boldsymbol{\mathrm{K}}$ corresponds to the intrinsic camera matrix. Given a 3D vector, the operator $(\cdot)^{\wedge}$ denotes its conversion to a skew-symmetric matrix. Furthermore, a vector with an overbar on it signifies its representation in homogeneous coordinates.
\begin{figure*}[t!]
    \centering
    \includegraphics[width=1\textwidth]{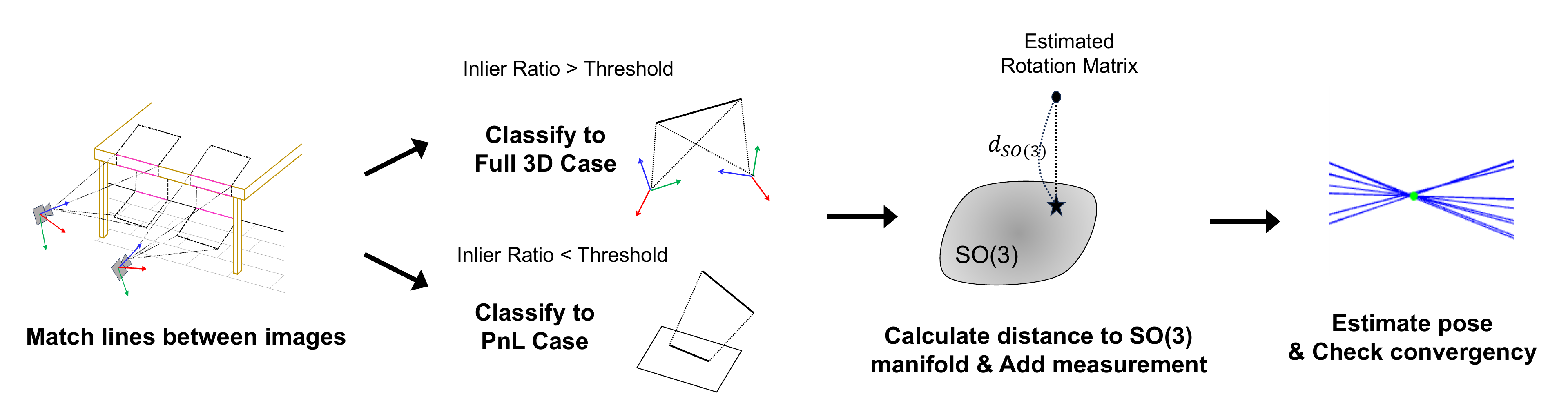}
    \caption{
    Pipeline of the proposed calibration algorithm. The procedure iteratively continues until the optimized solution reaches convergence and its associated cost drops below a predefined threshold.
    }
    \label{fig:pipeline}
    \vspace{-0.3cm}
\end{figure*}
\subsection{Merged Quadratic System}
\label{MQS}
To leverage 3D lines within our algorithm, we initiate by extracting and matching line features from a pair of RGB images using a novel deep matcher designed explicitly for line features \cite{Arxiv-2020-Pautrat}, enabling line matching between significantly different scenes as shown in \figref{fig:matching}. Based on the matching results, the 3D coordinates of the corresponding line are established via a line fitting method using \ac{RANSAC}, exploiting depth images. As depicted in \figref{fig:overview}, when the proportion of inliers exceeds a predetermined threshold in both images, the feature is categorized as the \( \textit{full 3D case} \). In this scenario, a ``3D point on 3D line'' constraint is incorporated into an equation system. On the other hand, if only the line associated with the source camera aligns accurately, the feature is classified as \( \textit{PnL case} \), and we apply a constraint as detailed in \cite{RAL-2020-Zhou}.

\subsubsection{Full 3D Case}

In the first scenario, we utilize a constraint that transformed 3D points on the source line should lie on the 3D target line. This representation begins from the property of dual Plücker matrix. Given a 3D point $\mathbf{X}$, the subsequent equation holds:
\begin{align}\begin{split}\label{equ:Full3D}
    \boldsymbol{\mathrm{L}}^*_t\bar{\boldsymbol{\mathrm{X}}}_{tj}=&\begin{pmatrix}
 -\boldsymbol{\mathrm{d}}^{\wedge}_t& \boldsymbol{\mathrm{m}}_t  \\
 -\boldsymbol{\mathrm{m}}_t^T& 0 \\
\end{pmatrix} \begin{pmatrix}
 \boldsymbol{\mathrm{R}} \boldsymbol{\mathrm{X}}_{sj} + \boldsymbol{\mathrm{t}}\\1
\end{pmatrix}\\[3pt]& =  \begin{pmatrix}
-\boldsymbol{\mathrm{d}}^{\wedge}_t \boldsymbol{\mathrm{R}}\boldsymbol{\mathrm{X}}_{sj} -\boldsymbol{\mathrm{d}}^{\wedge}_t\boldsymbol{\mathrm{t}}+\boldsymbol{\mathrm{m}}_t\\
-\boldsymbol{\mathrm{m}}_t^T\boldsymbol{\mathrm{R}}\boldsymbol{\mathrm{X}}_{sj}-\boldsymbol{\mathrm{m}}_t^T\boldsymbol{\mathrm{t}} \end{pmatrix}  \\[3pt]&= \begin{pmatrix}
 -\boldsymbol{\mathrm{d}}^{\wedge}_t \boldsymbol{\mathrm{R}}\boldsymbol{\mathrm{X}}_{sj}+ \boldsymbol{\mathrm{m}}_t\\-\boldsymbol{\mathrm{m}}_t^T\boldsymbol{\mathrm{R}}\boldsymbol{\mathrm{X}}_{sj}
\end{pmatrix} + \begin{pmatrix}
 -\boldsymbol{\mathrm{d}}^{\wedge}_t\boldsymbol{\mathrm{t}}\\
-\boldsymbol{\mathrm{m}}^{T}_t\boldsymbol{\mathrm{t}}\end{pmatrix} \\[3pt]&=\boldsymbol{0}_{4\times1}, \quad (j=1,2)
\end{split}
\end{align} where $j$ represents each endpoint, respectively. Thus, two constraints are derived from a single line pair $\boldsymbol{\mathrm{L}}_t \leftrightarrow \boldsymbol{\mathrm{L}}_s$.
\subsubsection{PnL Case} 
For the second case, we adopt the constraint, also utilized in \cite{RAL-2020-Zhou}. Since the projection of transformed 3D point $\boldsymbol{\mathrm{\hat{x}}}_{t}$ lies on 2D line $\boldsymbol{\mathrm{l}}_t$ of the target image, $\boldsymbol{\mathrm{l}}^T_t \boldsymbol{
\mathrm{\hat{x}}}_{tj} = 0$ holds. Consequently, from a singular correspondence \(\boldsymbol{\mathrm{l}}_t \leftrightarrow \boldsymbol{\mathrm{L}}_s\), we derive two constraints:
\begin{equation}\label{equ:PnL}
(\boldsymbol{\mathrm{K}}_t^T\boldsymbol{\mathrm{l}}_t)^T(\boldsymbol{\mathrm{RX}}_{sj}+\boldsymbol{\mathrm{t}}) = 0. \quad (j=1,2)
\end{equation}

\subsubsection{System Solver and Solution Refinement}
To merge \eqref{equ:Full3D} and \eqref{equ:PnL} into a unified system, we represent the rotation matrix $\boldsymbol{\mathrm{R}}$ using the CGR parameterization \cite{ICCV-2011-Mizaei}, defined as: 
\begin{equation}\label{equ:CGR}
\boldsymbol{\mathrm{R}}=\frac{\bar{\boldsymbol{\mathrm{R}}}}{1+\boldsymbol{\mathrm{s}}^T\boldsymbol{\mathrm{s}}}, \;\bar{\boldsymbol{\mathrm{R}}}=(1-\boldsymbol{\mathrm{s}}^T\boldsymbol{\mathrm{s}})\boldsymbol{\mathrm{I}}_3+2[\boldsymbol{\mathrm{s}}]_{\times}+2\boldsymbol{\mathrm{s}}^T\boldsymbol{\mathrm{s}},
\end{equation} where $\boldsymbol{\mathrm{s}}=[s_1, s_2, s_3]$. Multiplying $(1+\boldsymbol{\mathrm{s}}^T\boldsymbol{\mathrm{s}})$ to the both sides of \eqref{equ:Full3D} and \eqref{equ:PnL}, we get following equations which can be readily aggregated: 
\begin{align}\begin{split}\label{equ:CGRMultiplied}
 \begin{pmatrix}
 -\boldsymbol{\mathrm{d}}^{\wedge}_t \bar{\boldsymbol{\mathrm{R}}}\boldsymbol{\mathrm{X}}_{sj} +(1+\boldsymbol{\mathrm{s}}^T\boldsymbol{\mathrm{s}})\boldsymbol{\mathrm{m}}_t \\ -\boldsymbol{\mathrm{m}}_t^T\bar{\boldsymbol{\mathrm{R}}}\boldsymbol{\mathrm{X}}_{sj} 
\end{pmatrix} + & \begin{pmatrix}
 -\boldsymbol{\mathrm{d}}^{\wedge}_t\boldsymbol{\mathrm{\tau}}\\
-\boldsymbol{\mathrm{m}}^{T}_t\boldsymbol{\mathrm{\tau}}\end{pmatrix} =  \boldsymbol{0}_{4\times 1}, \\[3pt] (\boldsymbol{\mathrm{K}}_t^T\boldsymbol{\mathrm{l}}_t)^T\bar{\boldsymbol{\mathrm{R}}} \boldsymbol{\mathrm{X}} _{sj} + (\boldsymbol{\mathrm{K}}_t^T\boldsymbol{\mathrm{l}}_t)&\boldsymbol{\mathrm{\tau}} = 0, \quad (j=1,2),
\end{split}
\end{align} where $\boldsymbol{\mathbf{\tau}} = (1+\boldsymbol{\mathrm{s}}^T\boldsymbol{\mathrm{s}})\boldsymbol{\mathrm{t}}$. Given $M$ and $N$ measurements included in each case, we stack them vertically to have the following quadratic system:
\begin{equation}\label{equ:Quadratic}
\boldsymbol{\mathrm{Ar}}+\boldsymbol{\mathrm{B\tau}}=\boldsymbol{0}_{(8M+2N)\times1},
\end{equation} where  $\boldsymbol{\mathbf{r}} =[s_1^2,s_2^2,s_3^2,s_1s_2,s_1s_3,s_2s_3,s_1,s_2,s_3,1]^T$. Subsequently, \eqref{equ:Quadratic} is solved using the technique presented in \cite{RAL-2020-Zhou} to retrieve the rotation matrix and translation vector. Notably, the difference is that the system based on the CGR parameter is processed using the RE3Q3 solver \cite{ACCV-2019-Zhou}.

The system's initial value is refined by optimizing a cost function combining the line reprojection error $\boldsymbol{\mathrm{e}}_{l}$, as described in \cite{IROS-2017-Zuo}, and the 3D point-to-line error $\boldsymbol{\mathrm{e}}_{L}$ \cite{IROS-2020-Zhou}:

\begin{equation}\label{equ:LineReproj}
    \boldsymbol{\mathrm{e}}_{l} = \begin{bmatrix}
\frac{\boldsymbol{\mathrm{x}}_{t1}^T \hat{\boldsymbol{\mathrm{l}}}}{\sqrt{\hat{l}^2_1+\hat{l}^2_2}} & \frac{\boldsymbol{\mathrm{x}}_{t2}^T \hat{\boldsymbol{\mathrm{l}}}}{\sqrt{\hat{l}^2_1+\hat{l}^2_2}} \\
\end{bmatrix}^T,
\end{equation}
\begin{equation}\label{equ:3DPointonLine}
\boldsymbol{\mathrm{e}}_{L} = (\boldsymbol{\mathrm{I}}_3 - \boldsymbol{\mathrm{d}}_t\boldsymbol{\mathrm{d}}_t^T)(\boldsymbol{\mathrm{R}}\boldsymbol{\mathrm{X}}_{sj}+\boldsymbol{\mathrm{t}} - \boldsymbol{\mathrm{X}}_{tj}), \quad (j=1,2)
\end{equation}
 where $\hat{\boldsymbol{\mathrm{l}}}$ is the projected line on the target image. Using \eqref{equ:LineReproj} and \eqref{equ:3DPointonLine}, we define our cost function as follows:
\begin{equation}\label{equ:Optimization}
    \min_{\boldsymbol{\mathrm{R}},\boldsymbol{\mathrm{t}}}(\sum_{i=1}^{M}\boldsymbol{\mathrm{e}}_{L_i}^T\boldsymbol{\mathrm{e}}_{L_i} + \sum_{j=1}^{N}\boldsymbol{\mathrm{e}}_{l_j}^T\boldsymbol{\mathrm{e}}_{l_j}).
\end{equation} We employ Levenberg-Marquardt (LM) algorithm for optimization to refine the initial estimates of the parameters iteratively.

\subsection{Scene Selection}

This section introduces algorithms to identify informative lines for estimating the rotation matrix and translation. It is necessary to assess whether the resulting measurement provides sufficient information for an accurate estimation. Determining the rotation matrix by solving the quadratic system as section \ref{MQS} can be computationally demanding, considering that PeLiCal incorporates real-time features. Thus, our algorithm employs a lightweight approach based on a linear equation directly derived from the direction of the 3D line. Additionally, we evaluate the presence of a translation vector corresponding to the optimized rotation matrix in the form of a 3-dimensional line in the $\mathbb{R}^3$ space.

\subsubsection{Selecting Lines for Rotation Estimation}
\label{subsub:LineAdding}

For the $full \ 3D\ case$, we derive the following constraint relating the direction vectors from each camera:

\begin{equation}\label{equ_3d3drot}
    \boldsymbol{\mathrm{d}}_t = \boldsymbol{\mathrm{R}}\boldsymbol{\mathrm{d}}_s.
\end{equation} In another case, we establish an additional constraint ensuring that the preimage of the 2D target line should contain the transformed 3D source line:
\begin{equation}\label{equ_2d3drot}
    (\boldsymbol{\mathrm{P}}_t^T\boldsymbol{\mathrm{l}}_t)^T_{1:3}\boldsymbol{\mathrm{R}}\boldsymbol{\mathrm{d}}_s = 0,
\end{equation} where $\boldsymbol{\mathrm{P}}$ represents the projection matrix of the target camera. Incorporating both equations, we derive the following linear equation:
\begin{equation}\label{equ:RotCert}
    \boldsymbol{\mathrm{C}} \mathrm{vec}(\boldsymbol{\mathrm{R}}) = \boldsymbol{\mathrm{b}},
\end{equation} where $\boldsymbol{\mathrm{C}}$ is $(3M+N)\times \ 9$ matrix and $\boldsymbol{\mathrm{b}}$ is $(3M+N)$-dimensional vector derived by stacking multiple equations of \eqref{equ_3d3drot}, \eqref{equ_2d3drot}. After solving the equation and obtaining a 9-dimensional vector as a solution, we reshape it and attain a square matrix denoted as $\boldsymbol{\mathrm{M}}$ whose Singular Value Decomposition (SVD) is $\boldsymbol{\mathrm{U \Sigma V}}^T$. When a new line measurement is introduced, we solve \eqref{equ:RotCert}, subsequently projecting the resulting matrix onto the $SO(3)$ manifold using a special orthogonalization technique as detailed in \cite{NeurIPS-2020-Levinson}. This projected matrix is denoted as $\boldsymbol{\mathrm{R}}'$ and can be expressed as $\boldsymbol{\mathrm{U\Sigma' V}}^T$, where $\boldsymbol{\Sigma}'$ is a diagonal matrix with diagonal elements $(1,1,\mathrm{det}(\boldsymbol{\mathrm{UV}}^T))$. 

Assume we define a set of matrices, each having an SVD representation $\boldsymbol{\mathrm{U D V}}^T$, with $\boldsymbol{\mathrm{D}}$ being an arbitrary diagonal matrix, then, it can be proved that the minimum distance between any two elements within this set is equivalent to the Frobenius norm of the difference between the two diagonal matrices. Consequently, the distance between two matrices, $\mathbf{M}$ and $\boldsymbol{\mathrm{R}}'$, is computed as follows:
\begin{equation}
    \mathrm{d}_{SO(3)} = \left\| \Sigma - \Sigma' \right\|_{F},
\end{equation} where $\left\| \cdot \right\|_{F}$ means Frobenius norm. In the calibration process, only a new line pair that decreases the $\mathrm{d}_{SO(3)}$ distance is integrated.
\begin{algorithm}[t!]
   \begin{algorithmic}[1]
    \caption{Convergence Voting.}\label{euclid}
    \State  \textbf{Input:} $\epsilon_d$ and $\mathrm{l}_1$, $\cdots$, $\mathrm{l}_N$ from (14), (15); 
    \State  \textbf{Output:} Converged inliers;
    \State \% Compute equidistance points
    \State $\boldsymbol{m} \ = \mathrm{EquiDistance}$(\{$\mathrm{l}_1$, $\cdots$, $\mathrm{l}_N$\})  
    \State \% Initialize maximum inlier set
    \State $\mathcal{I} = \emptyset $
    \State \% Voting
          \For{$i = 1, \cdots, {N \choose 2}$}
        \State $\mathcal{K}_i \ = \ \emptyset$
        \For{$j = 1, \cdots, N$}
        \If{PointToLineDistance$(\boldsymbol{m}_i$, $\mathrm{l}_j) < \epsilon_d$ }
        \State $\mathcal{K}_i = \mathcal{K}_i \cup \{ j \}$ \% add to inlier set
        \EndIf
        \EndFor
        \If{$\mathrm{n}(\mathcal{K}_i) > \mathrm{n}(\mathcal{I}$)} \% update maximum inlier set
        \State $\mathcal{I} = \mathcal{K}_i $
        \EndIf
      \EndFor
    \State \textbf{return} $\mathcal{I}$
    \end{algorithmic}
\end{algorithm}
\subsubsection{Evaluating Existence of Translation} Despite exclusively employing measurements that bring the estimation closer to the $SO(3)$ manifold, it remains necessary to validate whether a translation vector can be derived from it. Our algorithm uses geometric conditions derived from the transformation of Plücker coordinates, distinct from those employed in constructing the quadratic system.

In $full\ 3D\ case$, by transforming the Plücker coordinates of the source line, we obtain the following equation:
\begin{align}
    \begin{split}
        \boldsymbol{\mathrm{m}}_t & = \boldsymbol{\mathrm{R}}\boldsymbol{\mathrm{m}}_s + \boldsymbol{\mathrm{t}}^{\wedge}\boldsymbol{\mathrm{R}}\boldsymbol{\mathrm{d}}_s \\& = \boldsymbol{\mathrm{R}}\boldsymbol{\mathrm{m}}_s - (\boldsymbol{\mathrm{R}}\boldsymbol{\mathrm{d}}_s)^{\wedge}\boldsymbol{\mathrm{t}}, 
    \end{split}
\end{align} where a solution is represented as:
\begin{equation}\label{equ:CertTransFull}
    \mathbf{t} = (\boldsymbol{\mathrm{R}}\boldsymbol{\mathrm{m}}_s-\boldsymbol{\mathrm{m}}_t)\times\boldsymbol{\mathrm{R}}\boldsymbol{\mathrm{d}}_s + k\boldsymbol{\mathrm{R}}\boldsymbol{\mathrm{d}}_s, 
\end{equation} with $k$ as an arbitrary scalar. This implies that the potential candidate for the translation vector should be situated along a 3D line, as expressed in \eqref{equ:CertTransFull}. In $PnL\ case$, when projecting the transformed line represented with Plücker coordinates, we derive the following constraint by multiplying two endpoints of the 2D target line by it:
\begin{equation}\label{equ:CertTransPnL}
([\boldsymbol{\mathrm{Rd}}_s ]_{\times}\boldsymbol{\mathcal{K}}^T\bar{\boldsymbol{\mathrm{x}}}_j)^T\boldsymbol{\mathrm{t}}= \bar{\boldsymbol{\mathrm{x}}}_j^T \boldsymbol{\mathcal{K}}\boldsymbol{\mathrm{Rm}}_s. \quad (j=1,2)
\end{equation}

For a single endpoint, \eqref{equ:CertTransPnL} provides a potential candidate for the translation vector within a 3D plane. Exploiting both endpoints, we obtain a 3D line since the solution should lie on the intersection of two planes. Similarly to the previous case, given a single pair of $PnL \ case$, the potential solution for the translation vector also conforms to a 3D line. It is surprise to note that, in a noiseless scenario where an accurate rotation matrix is known, if we can acquire two measurements from either of the two cases, the translation vector can be precisely determined as the point of intersection between the two lines.

\subsection{Outlier Rejection via Convergence Voting}

\begin{figure}[!b]
    \centering
    \includegraphics[width=1\columnwidth]{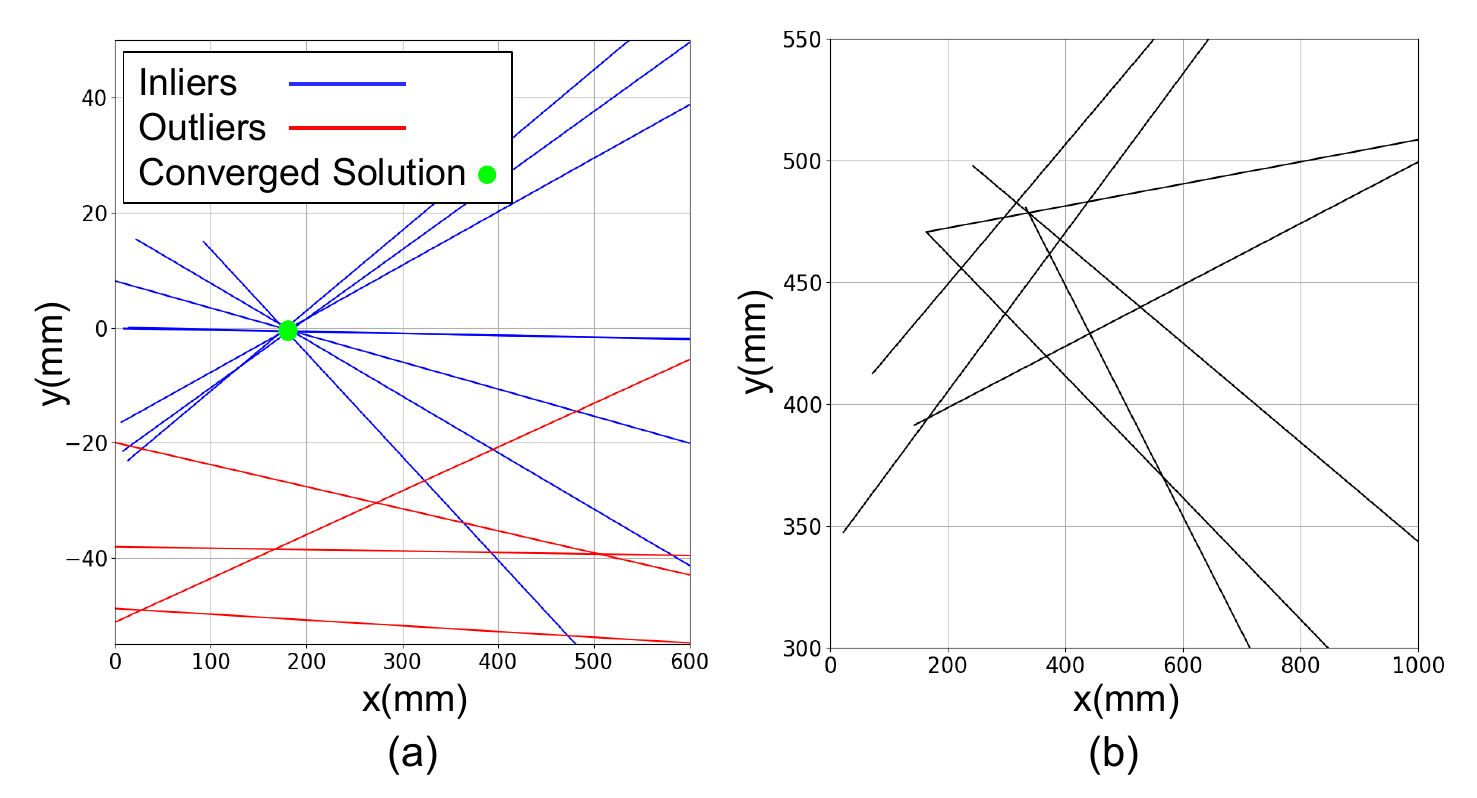}
    \caption{
    The results of the convergence voting process with candidate lines projected onto the $xy$-plane: (a) The algorithm successfully computes a convergence point, distinguishing inliers from outliers. (b) The candidate lines for translation do not converge due to an inaccurate rotation matrix or errors in the measurement.
    }
    \label{fig:voting}
    \vspace{-0.5cm}
\end{figure}

\begin{figure}[!t]
    \centering
    \includegraphics[width=1\columnwidth]{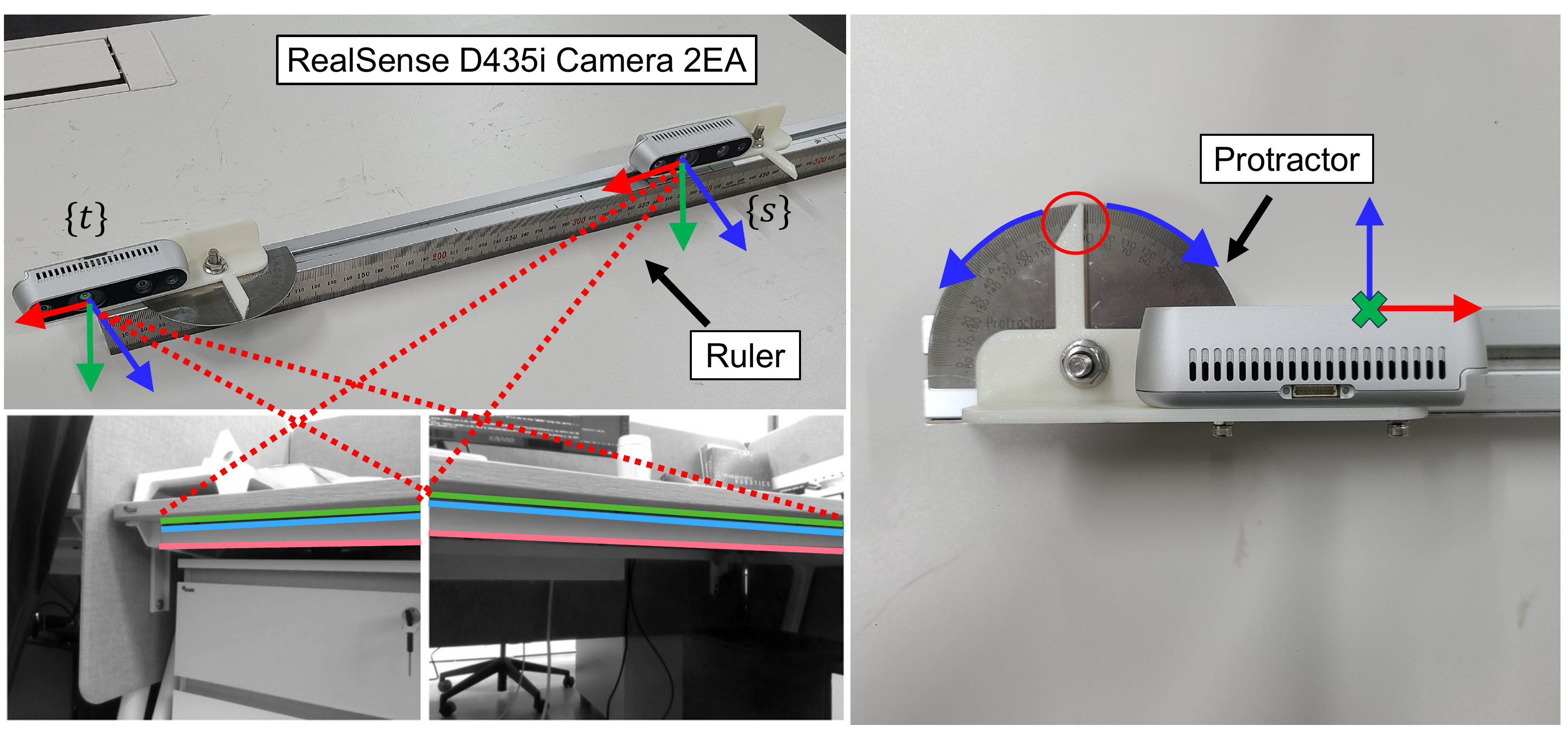}
    \caption{
    Equipment for accurate variation of rotation and translation between cameras. In the calibration process of our algorithm, the edge surface of the desk was used as a penetrating line.
    }
    \label{fig:device}
    \vspace{-0.5cm}
\end{figure}
In practical applications, noise in the line feature measurements diminish the precision of the rotation matrix. In such cases, we can still verify whether the estimated rotation matrix is sufficiently accurate to enable the existence of a translation vector as the convergence point of the candidate lines. To robustly determine the presence of a solution, we propose a ``\textit{convergence voting}'' algorithm with the pseudocode presented in Algorithm 1.

The algorithm starts by calculating equidistant points $\boldsymbol{\mathrm{m}}_i$ for every pair of provided candidate lines. These equidistant points serve as potential solutions for the translation vector. From $\boldsymbol{\mathrm{m}}_1$ to $\boldsymbol{\mathrm{m}}_N$, for each equidistant point, a set $\mathcal{K}_i$ consisting of lines located within a certain distance $\epsilon_d$ is defined. After obtaining the set for all points, if the number of elements in the maximum inlier set $\mathcal{I}$ surpasses a predefined threshold, the translation vector is considered to converge as depicted in \figref{fig:voting}. Then, we determine whether to finish the calibration depending on the residual of the cost function \eqref{equ:Optimization}. If the residual is not small enough, only the line pair included in the $\mathcal{I}$ is incorporated into the quadratic system. The algorithm repeats the process until two conditions are satisfied, as shown in \figref{fig:pipeline}.

\section{experiment}
\label{sec:experiment}

\subsection{Evaluation Metrics}
Our experiments were conducted on a setup running Ubuntu 20.04, powered by an Intel Core i7-12700@2.1 GHz CPU and an NVIDIA GeForce RTX 3080 GPU. We performed a two-fold evaluation to demonstrate the effectiveness of PeLiCal. The first evaluation focused on identifying subtle changes in pose by using an experimental setup as illustrated in \figref{fig:device}. In the next phase, we benchmarked the precision of PeLiCal against other algorithms in a stereo setting.

We compared our approach with three calibration methods: Kalibr \cite{ICRA-2016-Rehder}, ROS Calibrator, which is re-implementation of \cite{TPAMI-2000-Zhang} and CamMap \cite{RAL-2022-Xu}, an extrinsic calibration strategy designed to align maps formulated by ORB-SLAM3 \cite{TRO-2021-Campos}. In real-world settings, obtaining a definitive ground truth for calibration is often challenging. As a result, our initial experiment aimed to evaluate the accuracy of the pose changes detected by our algorithm, avoiding the need for such ground truth. Likewise, due to this inherent challenge, our comparison against other techniques was based on the criteria proposed by Perez et al. \cite{RAL-2017-Perez}.
\subsubsection{Verification of Pose Variance Estimation}

Each of the RealSense D435i cameras is mounted on an aluminum structure in our device. The target camera features a protractor to assess rotational movements, while the source camera can be moved along the x-axis, determining translational changes with the ruler in place. While manual adjustments to the angle and spacing of the cameras may not result in the exact anticipated pose disparity, a precise algorithm should be able to discern and quantify the difference within a reasonable error margin.

From a base setup where both cameras faced forward with a \unit{20}{cm} separation, we evaluated poses in 30 different scenarios. Within these tests, the angles varied between 0\textdegree\xspace and 80\textdegree\xspace, increasing in 20\textdegree\xspace steps, and the distances extended from \unit{20}{cm} to \unit{45}{cm}, with a progression of \unit{5}{cm} at each step. Following this, we analyzed the accuracy of a \unit{5}{cm} translational change along the x-axis at a constant angle and an angular adjustment of 20\textdegree\xspace at a set distance.
The error for the $i$th rotation or translation, keeping the alternate parameter unvarying at either $\mathrm{X}cm$ or $\theta^{\circ}$, is computed as follows:
\begin{align}
    \begin{split}
 & \theta^{\mathrm{X}cm}_{e_i} = \left\| (\boldsymbol{\theta}^{\mathrm{X}cm}_{i+1} - \boldsymbol{\theta}^{\mathrm{X}cm}_i) - (0,20,0)^T \right\|_2, \\[2pt]
& \mathrm{X}^{\theta^{\circ}}_{e_i} =\left| \left\| \mathbf{X}^{\theta^{\circ}}_{i+1}  - \mathbf{X}^{\theta^{\circ}}_{i}\right\|_2 -5 \right|.
    \end{split}
\end{align} 

\subsubsection{Performance Comparison by Checkerboard}
\begin{figure}[t!]
    \centering
    \includegraphics[width=1\columnwidth]{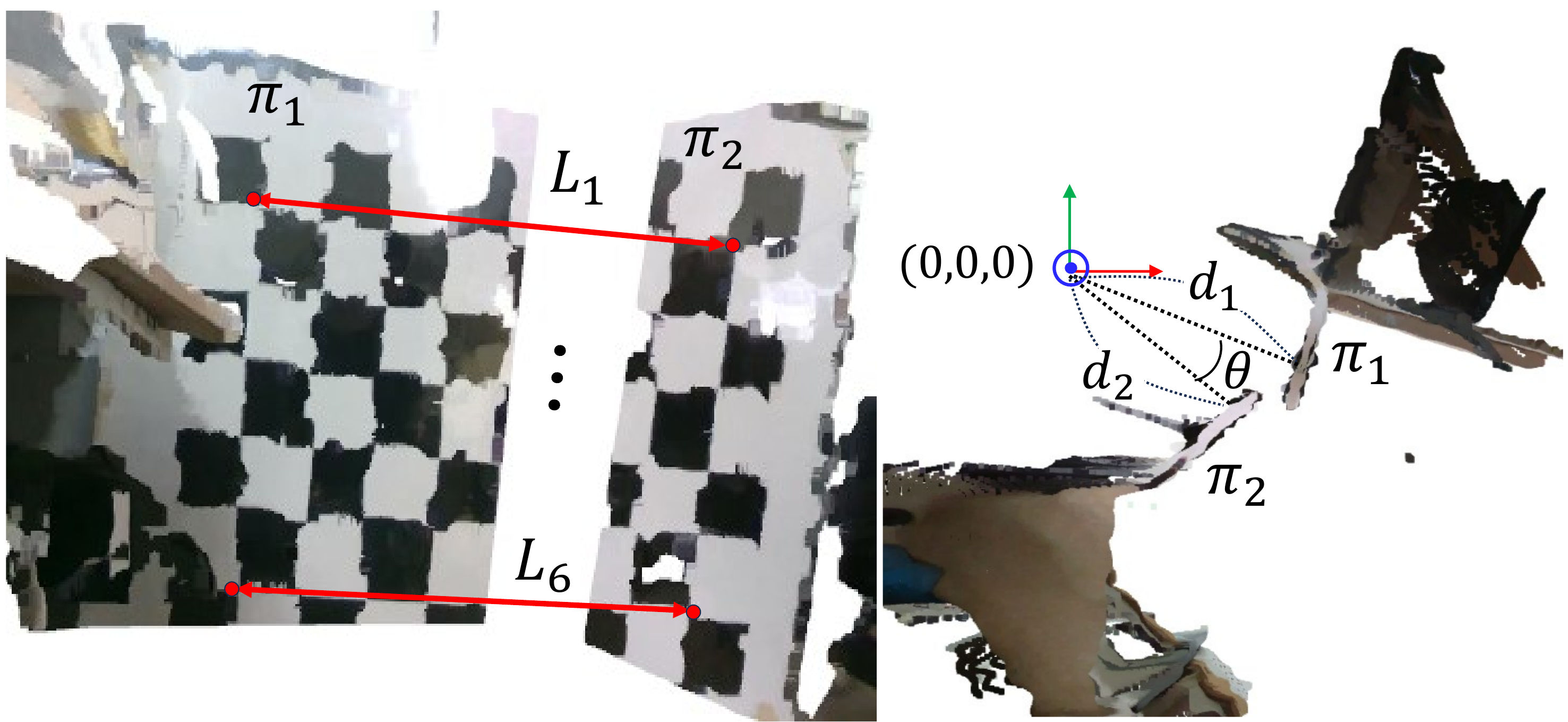}
    \caption{
The reconstructed point cloud representing the checkerboard plane, denoted as \(\pi_1\) and \(\pi_2\), is merged into a singular coordinate using the predicted pose. The value \( l \) is derived by averaging the ratios of \( L_1 \) to \( L_6 \) for the number of squares in a row, then determining the discrepancy from the actual length of \unit{108}{mm} (left). The variable \( d \) denotes the disparity between \( d_1 \) and \( d_2 \), which correspond to the distances of \(\pi_1\) and \(\pi_2\) from the origin, respectively. Meanwhile, \( \theta \) represents the angle formed between the normals of the two planes (right).}
\label{fig:metrics}
\vspace{-0.5cm}
\end{figure}

First, we recorded a scene where a sizable calibration board was visible to both cameras. From each snapshot, we isolated planes, transformed them into 3D spatial coordinates, and combined them into one coordinate system using external parameters. Afterward, we identified the most distant corner points. By dividing the span between these corners by the count of squares, we derived $l$, representing the variance between the mean square dimension of the board and its actual size. An ideal estimation brings this value closer to zero. Moreover, we examined the deviations in the distance from the origin, denoted as $d$, and the angular disparity between the plane normals, represented by $\theta$. If mapped precisely, both metrics should ideally converge to zero, implying perfect alignment of the two planes. Detailed illustration of the metrics is provided in \figref{fig:metrics}.

\begin{figure*}[!t]
\begin{minipage}{.67\textwidth} %
  \centering
  \subfigure[Error Measured for Rotation (\textdegree) and Translation (\cm) Variation]{%
  \resizebox{0.92\textwidth}{!}{
  \begin{tabular}{c|cccc|c|ccccc}
    \toprule[1.5pt] 
    \midrule[0.5pt]
    Fixed & \multicolumn{4}{c|}{Rotation Variation (\textdegree)} &  Fixed & \multicolumn{5}{c} {Distance Variation (\cm)} \\
    Distance & 0$\rightarrow$20 & 20$\rightarrow$40 & 40$\rightarrow$60 & 60$\rightarrow$80 & Angle & 20$\rightarrow$25 & 25$\rightarrow$30 & 30$\rightarrow$35 & 35$\rightarrow$40 & 40$\rightarrow$45
    \\ \midrule[1pt]
    \unit{45}{cm} & 0.6398  & 0.0707 & 0.9675 & 1.2077 & 80\textdegree & 0.1697 & 0.1137 & 1.3918 &0.6780 & 1.3741 \\
    \unit{40}{cm} & 0.7586  & 0.1131 & 0.3187 & 0.2768 & 60\textdegree & 0.1584 & 1.0096 & 1.1799 & 1.3456 & 0.4132 \\
    \unit{35}{cm} & 0.1203  & 0.4511 & 0.8060 & 0.1947 & 40\textdegree & 0.3156 & 0.1831 & 0.7042 & 0.6292 & 0.5332 \\
    \unit{30}{cm} & 0.3974 & 0.2406 & 0.2435 & 0.3669 & 20\textdegree & 0.0120 & 0.0093 &0.3852 & 0.1017 & 0.1975 \\
    \unit{25}{cm} & 0.1790  & 0.1944 & 0.4898 & 0.1901 &  0\textdegree & 0.2690 & 0.2260 & 0.1161 & 0.1105 & 0.0595
    
    \\
    \unit{20}{cm} & 0.4953  & 0.4800 & 0.4810 & 1.1286 & & & & & & \\
   \midrule[0.5pt]
   \bottomrule[1pt]
   \end{tabular}
  }
}
  
\end{minipage}%
\begin{minipage}{.37\textwidth} %
  \centering
  \subfigure[Error Heatmap]{%
    \includegraphics[height=0.12\textheight,width=0.9\columnwidth]{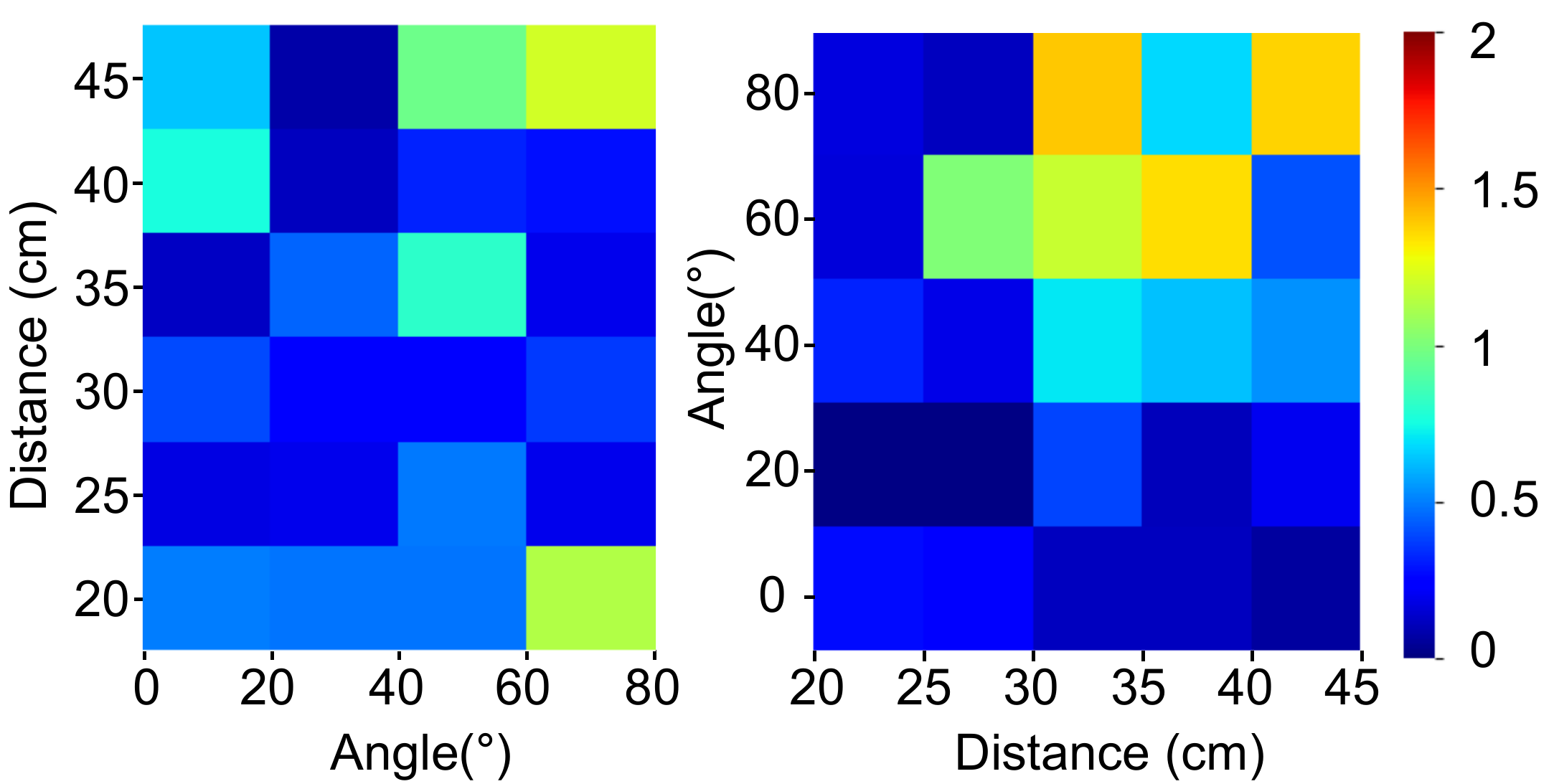}
    \label{fig:heatmap}
  }
\end{minipage}
\caption{Discrepancies between estimated pose variance and its actual value. As rotation and translation trend toward the top-right quadrant of the heatmap, error rates escalate. This elevation in error is primarily ascribed to a diminishing count of diverse feature matches as both cameras progressively capture different viewpoints.}
\label{fig:diff_fov}
\vspace{-0.3cm}
\end{figure*}

\subsection{Calibration with Various Common FOV}

The experiment outcomes are presented in the table with the heatmap in \figref{fig:diff_fov}. In setups characterized by short distances and minimal angles where a shared FOV between the cameras is significant, the outcomes are closely aligned with the actual modifications in both parameters. Notably, in the first column with the minimum variation for each scenario, we observed a translational error of up to \unit{0.3156}{cm} and an angular error of 0.7586\textdegree. However, as the angle exceeded 40\textdegree\xspace and the FOV no longer overlapped, the errors escalated due to the diminished number of line features. This is because as the angle between the camera's central axis and the line expands, the accuracy of the depth sensor readings diminishes. Nonetheless, even under these challenging conditions, where traditional calibration techniques that require specialized tools falter, PeLiCal showed discrepancies reaching up to 1.2077\textdegree\xspace and \unit{1.3918}{cm}.

\subsection{Stereo Calibration}

In stereo camera systems, an increase in the baseline reduces the overlapping \ac{FOV} between the cameras. Considering this, we evaluated the extrinsic calibration performance at baselines of \unit{30}{cm} (near) and \unit{45}{cm} (far) to examine the effects of a diminished shared \ac{FOV}. For calibration purposes, Kalibr utilized an A3-sized AprilTag, while the ROS Calibrator employed an $8\times5$ checkerboard with squares measuring \unit{45}{mm} each. We also tested CamMap in visually feature-rich indoor environments, such as labs or offices, selecting the setup that produced the the minimal errors.

\begin{figure}[!b]
    \centering
    \includegraphics[width=1\columnwidth]{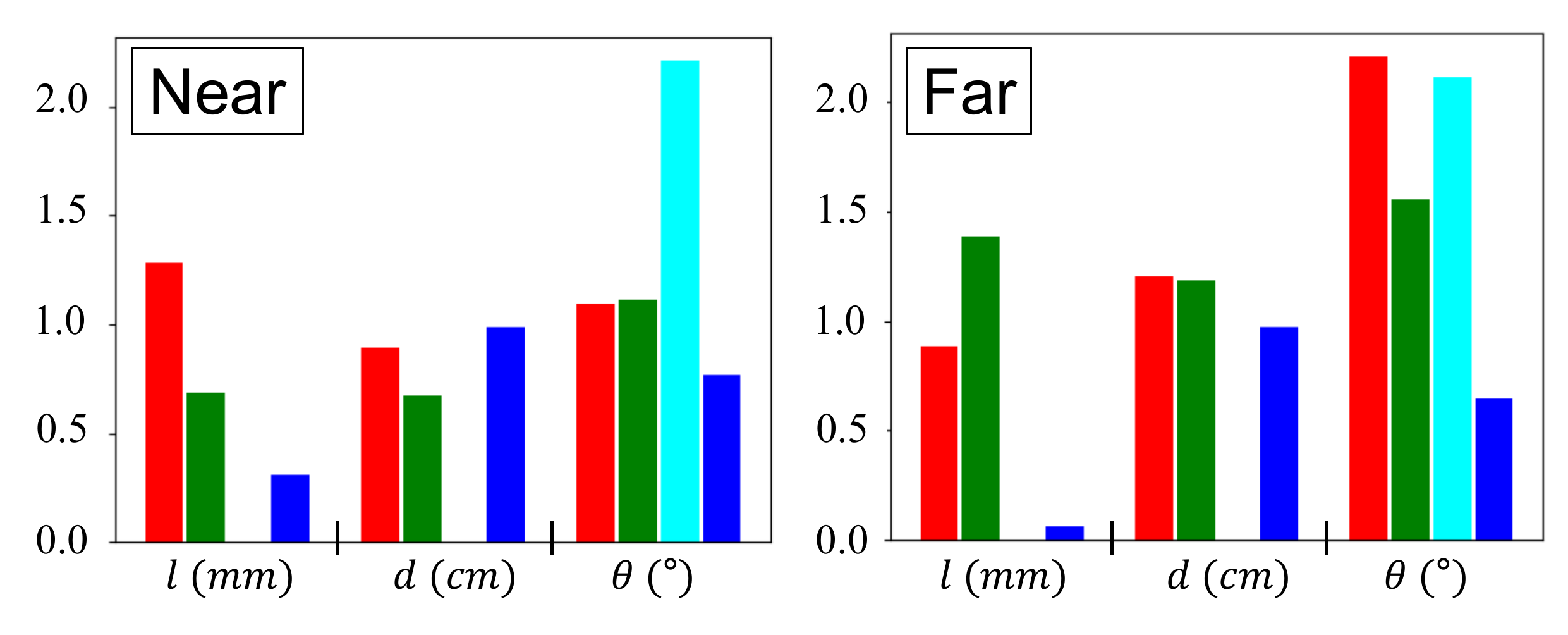}
    \caption{ Experimental results of stereo calibration  (30cm (left), 45cm (right))  for each algorithm (Kalibr (red), ROS Calibrator (green), CamMap (cyan), PeLiCal (blue)).
    }
\label{fig:graph}
\vspace{-5mm}
\end{figure}

The result is shown in \tabref{demo-table} with \figref{fig:graph}. The best results for each metric are emphasized in bold. In the first configuration, the ROS Calibrator most accurately determined the distance difference of the two planes from the origin. For the remaining metrics and configuration, PeLiCal predominantly outperformed others. Although CamMap effectively produced its trajectory and map over multiple sequences, it fails to estimate the given translational configuration. As a result, its associated values were not incorporated in the table. Also, while the accuracy of the ROS Calibrator and Kalibr diminished with the reduction of common FOV caused by a more increased baseline, our algorithm consistently maintained its accuracy without substantial deviation, attributing to its utilization of extended, distinct line features spanning across the images for calibration.


\section{Conclusion}
\label{sec:conclusion}
\begin{table}[!b]
\vspace{-3mm}
\renewcommand{\arraystretch}{1.5}
\centering
\caption{\label{demo-table}Estimated rotation $\Delta\mathbf{R}$ (roll, pitch, yaw), translation $\Delta\mathbf{t}$ ($x,y,z$) and metrics by merging two planes ($l$, $d$, $\theta$) for each algorithm in two configurations. The results with substantial errors are replaced with a hyphen (---). }
    \resizebox{0.48\textwidth}{!}{%
        \begin{tabular}{|cl|c|c}
        \toprule[1.5pt] 
        \midrule[0.5pt]
        \multicolumn{2}{c|}{Baseline}                          & Near             & Far                        \\ 
        \midrule
        \multicolumn{1}{c|}{\multirow{3}{*}{\rotatebox{90}{\texttt{Kalibr}}}}  & $\Delta\mathbf{R}$ (\textdegree) & {[}0.66, 0.08, -0.01{]}   & {[}0.22, 0.98, 0.18{]}   \\
        \multicolumn{1}{c|}{}                                  & $\Delta\mathbf{t}$ (\mm)    & {[}296.46, 2.47, 1.36{]} & {[}441.94, 3.12, -7.96{]}  \\
        \multicolumn{1}{c|}{}                                  & [$l$, $d$, $\theta$] & {[}1.28, 0.89, 1.09{]}  & {[}0.88, 1.21, 2.21{]}   \\ \hline
        \multicolumn{1}{c|}{\multirow{3}{*}{\rotatebox{90}{\begin{tabular}[c]{@{}c@{}} \texttt{ROS}\end{tabular}}}}     & $\Delta\mathbf{R}$ (\textdegree)       & {[}-0.56, 0.29, -0.08{]}    & {[}-0.24, 0.55, -0.15{]}     \\
        \multicolumn{1}{c|}{}                         & $\Delta\mathbf{t}$ (\mm) & {[}297.79, -4.48, -2.46{]} & {[}451.54, -7.50, 0.00{]} \\
        \multicolumn{1}{c|}{}                         & [$l$, $d$, $\theta$] & {[}0.69, \textbf{0.67}, 1.11{]}   & {[}1.39, 1.19, 1.55{]}   \\ \hline
        \multicolumn{1}{c|}{\multirow{3}{*}{\rotatebox{90}{\texttt{CamMap}}}}  & $\Delta\mathbf{R}$ (\textdegree)       & {[}0.04, -0.19, 1.10{]}   & {[}2.23, 0.15, -0.47{]}    \\
        \multicolumn{1}{c|}{}                         & $\Delta\mathbf{t}$ (\mm) & ---                          & ---                         \\
        \multicolumn{1}{c|}{}                         & [$l$, $d$, $\theta$] & {[} --- , --- , 2.21{]}            & {[} --- , --- , 2.11{]}             \\ \hline
        \multicolumn{1}{c|}{\multirow{3}{*}{\rotatebox{90}{\texttt{PeLiCal}}}} & $\Delta\mathbf{R}$ (\textdegree)       & {[}0.00, 0.10, 0.00{]}   & {[}0.00, -0.23, 0.00{]}    \\
        \multicolumn{1}{c|}{}                         & $\Delta\mathbf{t}$ (\mm) & {[}303.00, -3.14, 2.02{]} & {[}453.63, -5.33, 3.76{]}   \\
        \multicolumn{1}{c|}{}                         & [$l$, $d$, $\theta$] & {[}\textbf{0.31}, 0.98, \textbf{0.77}{]}  & {[}\textbf{0.07}, \textbf{0.97}, \textbf{0.65}{]}   \\
        
        \midrule[0.5pt]
        \bottomrule[1pt]
        \end{tabular}%
    }

\end{table}

We present a novel extrinsic calibration algorithm for RGB-D cameras leveraging line features. Our method solves the merged quadratic system derived from two conditions for initial pose estimation. Moreover, exploiting the constraints of transformed Plücker coordinates, translation candidates were represented as 3D lines. Existence of the optimal translation was validated through a convergence assessment through voting. The algorithm showed its ability to adapt to slight variations in pose, achieving performance superior to algorithms employing a calibration pattern.
\newpage


\balance
\small
\bibliographystyle{IEEEtranN} 
\bibliography{string-short,references}

\end{document}